\documentclass[letterpaper, 10 pt, conference]{ieeeconf}  
\usepackage{graphicx}  
\usepackage{amssymb}
\usepackage{siunitx}
\usepackage{makecell}
\usepackage{float}
\usepackage{hyperref}
\usepackage{caption}
\usepackage{makecell}
\usepackage{siunitx}
\hypersetup{
    colorlinks=true,
    linkcolor=blue,
    filecolor=magenta,      
    urlcolor=cyan,
    pdftitle={FORCES Framework},
    pdfpagemode=FullScreen,
}
\IEEEoverridecommandlockouts                              

\title{\LARGE \bf
A Framework for Closed-Loop Robotic Assembly, Alignment and Self-Recovery of Precision Optical Systems
}

\author{Seou Choi$^{1\ast}$, Sachin Vaidya$^{1,2\ast}$, Caio Silva$^{1}$, Shiekh Zia Uddin$^{3}$, \\ Sajib Biswas Shuvo$^{4}$, Shrish Choudhary$^{1}$, and Marin~Solja\v{c}i\'{c}$^{1,2}$ 
 \thanks{$^\ast$ These authors contributed equally.}
 \thanks{$^{1}$ Seou Choi, Sachin Vaidya, Caio Silva, Shrish Choudhary, and Marin~Solja\v{c}i\'{c} are with Research Laboratory of Electronics, Massachusetts Institute of Technology, Cambridge, MA 02139, USA {\tt\small\{seouc130, svaidya1, caiosiq, shrishc, soljacic\}@mit.edu}}
 \thanks{$^{2}$ Sachin Vaidya  and Marin~Solja\v{c}i\'{c} are with NSF Institute for Artificial Intelligence and Fundamental Interactions and the Department of Physics, Massachusetts Institute of Technology, Cambridge, MA 02139, USA}
 \thanks{$^{3}$ Shiekh Zia Uddin is with Nokia Bell Labs, Murray Hill, NJ 07974, USA {\tt\small szistiak@gmail.com}}
 \thanks{$^{4}$  Sajib Biswas Shuvo is with School of Electrical, Computing and Energy Engineering, Arizona State University, Tempe, AZ 85287, USA {\tt\small sshuvo13@asu.edu}}}
  
\begin{document}

\maketitle
\pagestyle{empty}

\begin{abstract}
Robotic automation has transformed scientific workflows in domains such as chemistry and materials science, yet free-space optics, which is a high-precision domain, remains largely manual. Optical systems impose strict spatial and angular tolerances, and their performance is governed by tightly coupled physical parameters, making generalizable automation particularly challenging. In this work, we present a robotics framework for the autonomous construction, alignment, and maintenance of precision optical systems. Our approach integrates a hierarchical computer vision system, optimization routines, and custom-built tools to achieve this functionality. As a representative demonstration, we perform the fully autonomous construction of a tabletop laser cavity from randomly distributed components. The system performs several tasks such as laser beam centering, spatial alignment of multiple beams, resonator alignment, laser mode selection, and self-recovery from induced misalignment and disturbances. By achieving closed-loop autonomy for highly sensitive optical systems, this work establishes a foundation for autonomous optical experiments for applications across scientific domains. Project link: \url{https://github.com/seouchoikr/AutomateOptics}

\end{abstract}

\section{INTRODUCTION}
Robotic automation is increasingly transforming scientific workflows across chemistry, materials science, and biology~\cite{coley2019robotic,zhang2025multimodal,angers2025roboculture, szymanski2023autonomous, macleod2020self, rapp2024self, kandel2023demonstration, tom2024self, darvish2025organa}. Autonomous synthesis platforms now execute complex reaction sequences, and robotic microscopy systems perform adaptive characterization at scale. These systems leverage advances in robotics, perception, planning, and feedback control to accelerate discovery while improving reproducibility. In highly structured laboratory environments, robotics enables the abstraction of experiments into programmable procedures, thereby reducing human-induced variability and allowing for iterative optimization at speeds that are difficult to achieve manually. Despite this progress, the degree of automation realized across disciplines remains uneven.

Optics represents one of the most foundational yet least automated experimental domains. Optical systems underpin a broad spectrum of scientific areas, including quantum information, atomic and molecular physics, materials characterization and spectroscopy, microscopy, and astrophysics~\cite{northup2014quantum,balasubramanian2023imagining,kaiser2024evolvement}. Furthermore, stable optical setups are essential for industrial applications such as laser fabrication systems, biomedical imaging, semiconductor manufacturing, optical communication, and prototyping of commercial photonic devices, all of which make automation highly desirable. 

In laboratories, optical setups are meticulously assembled to explore new configurations, measurements, and hypotheses. However, optical experimentation is a high-precision domain, governed by stringent spatial and angular tolerances that couple directly to physical observables. A displacement of tens to hundreds of micrometers or a sub-degree angular deviation of a component can render an optical system completely unusable. Furthermore, optics is inherently reconfigurable and heterogeneous: components differ widely in geometry, sensitivity, and required precision. These characteristics make generalizable automation in optics particularly challenging. Successful automation must integrate perception, sub-millimeter manipulation, real-time feedback, and iterative closed-loop optimization rather than relying solely on simple pick-and-place operations.

In this work, we address these challenges through a robotics framework designed specifically for the assembly, alignment and maintenance of complex and arbitrary optical systems. Our approach integrates minimal but meaningful laboratory structuring, hierarchical computer vision and manipulation systems, and custom tools to achieve the required precision for closed-loop control over beam propagation and alignment. As a representative demonstration of the platform’s capabilities, we present the autonomous construction, alignment, and maintenance of a tabletop laser cavity, starting from scratch, i.e., from randomly distributed optical components in the laboratory workspace. The system performs spatial alignment of cavity elements, angular optimization to establish resonator stability, mode selection, and automated recovery from induced misalignment or external disturbance. By achieving closed-loop autonomy on a task that is traditionally regarded as highly sensitive and expertise-driven, this work takes a concrete step towards fully autonomous optical laboratories.

\section{RELATED WORKS}
\subsection{Automation platform for free-space optics}
\begin{figure*}[htbp]
\centering
\includegraphics[width=0.95\textwidth]{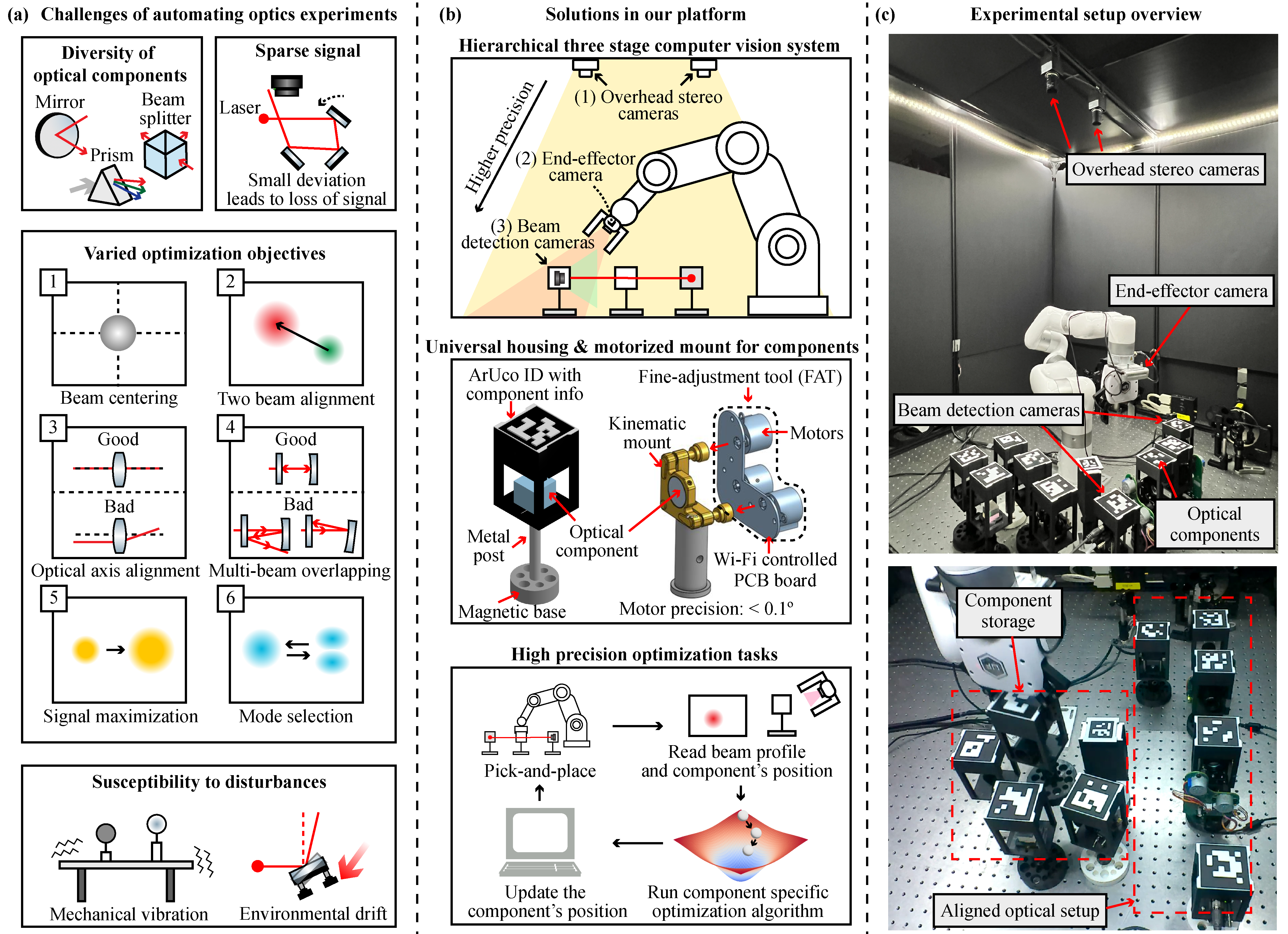}
\caption{A general overview of the \textbf{(a)} challenges associated with automating reconfigurable optics experiments, \textbf{(b)} the solutions proposed and implemented in this work, and \textbf{(c)} experimental implementation in a standard optics laboratory.}
\label{fig:challenge_and_solution}
\end{figure*}

In our previous work, we demonstrated AI-driven robotics for optics, introducing a general-purpose framework integrating generative AI, computer vision, and robotic manipulation to automate the optics experimental loop~\cite{uddin2025ai}. This work established a modular platform consisting of fiducial-based component identification, computer vision for pose estimation, and a robotic arm for pick-and-place operations. Our previous platform successfully demonstrated autonomous assembly of simple optical systems such as Michelson interferometers and imaging setups, as well as automated beam characterization and spectroscopy. Nevertheless, robotic manipulation was limited to pick-and-place of optical components and simple alignment procedures, often performed in isolation from real-time optical feedback. 

Building on this foundation, here we employ a feedback-driven strategy to support the full automation of high-precision optical systems, tackling all major challenges encountered in experimental optics (discussed in section III). This is achieved along three critical directions: (i) modularizing optics experiments into smaller automation tasks, (ii) formalizing optical alignment as an optimization problem, and (iii) actively monitoring and recovering optical setups under external perturbations. 

\subsection{Task-Specific Automation in Optics}
Automated alignment of optical systems has been studied in several narrow, task-specific contexts, particularly using reinforcement learning (RL) or model-based control. Sorokin et al. demonstrated alignment of an optical interferometer using an RL agent operating in a discrete action space~\cite{sorokin2020interferobot}. Makarenko et al. extended this framework to continuous action spaces and incorporated beam divergence control~\cite{makarenko2022aligning}. Fang and Savransky employed focal-plane sensing combined with Kalman filtering for automated alignment of reconfigurable optical systems~\cite{fang2016automated}. Mareev et al. and Wang et al. explored RL-based self-adjusting optical systems and beamline alignment, respectively~\cite{mareev2023self, wang2024action}, while Burkhardt et al. investigated active lens alignment using reinforcement learning~\cite{burkhardt2025active}. Additional contributions include model-free RL under noisy actuation~\cite{richtmann2024model}, deep RL for coherent beam combining~\cite{tunnermann2019deep}, and compact experimental platforms such as the Raspberry Pi auto-aligner~\cite{mathew2021raspberry}. Barwicz et al. demonstrated automated photonic packaging pipelines~\cite{barwicz2018automated}.

These works collectively demonstrate that learning-based control and automation can achieve alignment in narrow and repetitive optical tasks. In these works, the alignment problem is often reduced to tuning a small number of actuators within an already assembled system. In contrast, our framework enables generalizable autonomous construction and stabilization of setups at the system level. 

\section{PROBLEM DEFINITION}
Automating free-space optical experiments at the system level introduces challenges that differ fundamentally from conventional robotic manipulation tasks. Optical experiments are spatially reconfigurable, precision-constrained, and governed by tightly coupled physical observables. We identify four universal challenges that must be addressed to enable generalizable automation in optics (Fig.~\ref{fig:challenge_and_solution}(a)).

\textbf{Diversity of optical components:} Optical experiments consist of a heterogeneous collection of components such as mirrors, lenses and cameras, each with distinct geometries, optical functions, and alignment sensitivities. These components differ not only in size and shape but also in how misplacements affect system-level behavior. A robotic system must therefore operate across this diversity, e.g. with standardized handling, but without sacrificing the flexibility required for reconfigurable experimental layouts.

\textbf{Sparse signals:} Optical signals often exhibit highly nonlinear responses to small spatial or angular perturbations of components. Sub-millimeter translations or sub-degree rotations can extinguish the detected signals entirely. Furthermore, many measurable quantities yield sparse or intermittent feedback, where measurable signals only emerge within narrow regions of the components' parameter space.

\textbf{Varied optimization objectives:} Optical experiments also require diverse and context-dependent optimization criteria which may vary significantly between components within the same setup. These include centering beams along a predefined axis, aligning multiple beams, matching the optical axis to the beam path, achieving multi-beam spatial overlap, maximizing optical signal, or optimizing for specific optical spatial modes.

\textbf{Susceptibility to disturbances:} Free-space optical systems are intrinsically sensitive to environmental disturbances, including mechanical vibration and thermal drift. Minor perturbations can shift alignment beyond operational tolerances, rendering the setup unusable. In conventional laboratory practice, recovery is manual and reactive. Meaningful automation in optics therefore requires persistent monitoring and the ability to detect, diagnose, and correct deviations to maintain long-term stability of a setup.

\section{METHODS AND SOLUTIONS}
Here we present our robotics framework that combines structured hardware design, computer vision, and feedback-driven optimization. These aspects target all of the fundamental challenges in automating optics discussed above.

\begin{figure*}
\centering
\includegraphics[width=1.0\textwidth]{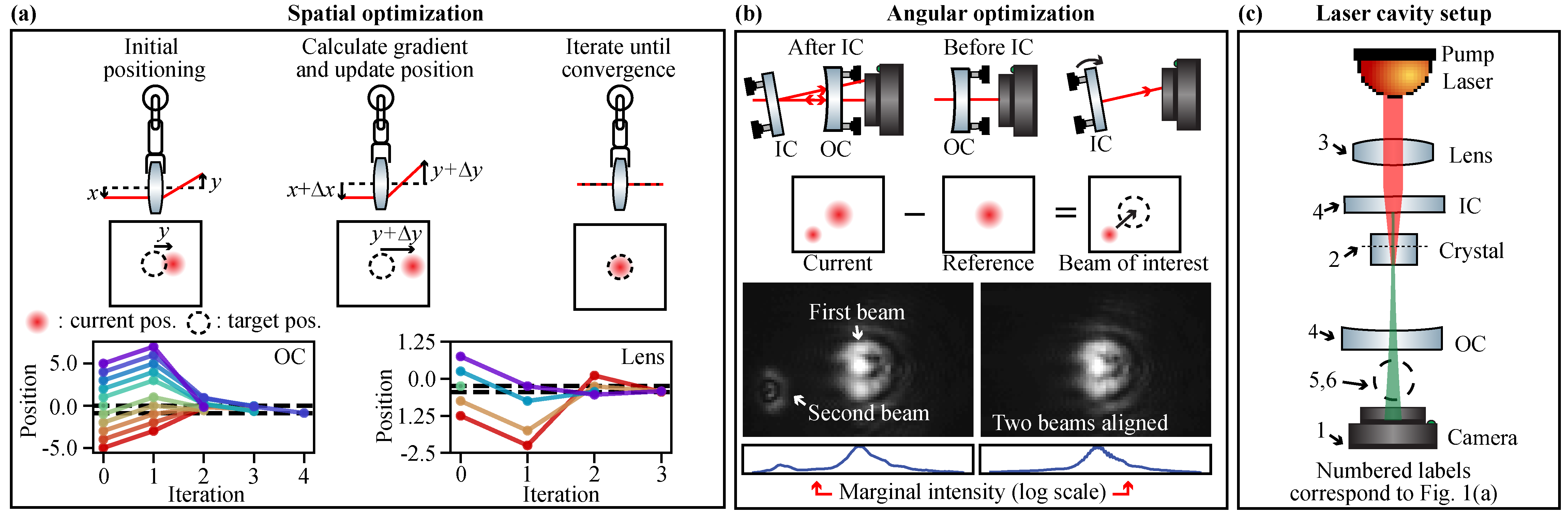}
\caption{High-precision automated optical experiments using \textbf{(a)} spatial optimization and \textbf{(b)} angular optimization. For the camera image in \textbf{(b)}, the original image was enhanced following the logarithmic intensity transformation: $I_{\textrm{processed}}= \textrm{log}(1+10I_{\textrm{raw}})/\textrm{log(1+10)}$, where $I_{\textrm{processed}}$ and $I_{\textrm{raw}}$ are the normalized intensities of the processed and raw images. \textbf{(c)} The fundamental architecture of a laser cavity setup and its associated alignment objectives. OC: out-coupler, IC: in-coupler. }
\label{fig:coarse_fine_adjustment}
\end{figure*}

\subsection{Overview of the platform}
\textbf{Robotic arm:} The core manipulation unit is a 7-degree-of-freedom (DOF) UFACTORY xArm7 robotic arm with a \SI{61}{\centi\meter} reach with a listed precision of \SI{\pm 0.1}{\milli\meter} and a payload capacity of \SI{3.5}{\kilo\gram}. 
The robotic arm is equipped with an end-effector that allows for gripping and handling of components (Fig.~\ref{fig:challenge_and_solution}(c)).

\textbf{Computer vision system:} To mitigate the sparse nature of optical signals, we employ a hierarchical computer vision system consisting of three subsystems: overhead stereo 4K cameras for coarse pose estimation and workspace monitoring, an end-effector LiDAR camera for accurate grasping, and beam detection cameras for optical feedback during spatial and angular optimization (Fig.~\ref{fig:challenge_and_solution}(b), top). 

\textbf{Modular optical component infrastructure}: For the reliable handling of a variety of optical components without sacrificing reconfigurability, we place all optical components (including ones with kinematic mounts) in custom 3D-printed housings with standardized geometry (Fig.~\ref{fig:challenge_and_solution}(b), middle). Each housing includes ArUco fiducial markers~\cite{garrido2014automatic} for identification and a magnetic base with embedded neodymium magnets and a rubberized bottom to mitigate slippage. The housings and bases are easy to implement and fully compatible with standard optics laboratories.

\textbf{Fine-adjustment tool (FAT):} To enable complex optimization routines involving many components, we develop a custom Wi-Fi controlled, motorized tool that engages with the fine-adjustment knobs located on the kinematic mounts of standard optical components. This tool consists of an Arduino Nano microcontroller and motor controllers integrated onto a printed circuit board (PCB), supporting up to three motors that can interface with standard optical mounts. The couplers that connect the motor shaft to the fine-adjustment knobs are 3D printed. This tool allows the system to remotely perform adjustments of specific components in combination with the pick-and-place operations of the robotic arm (Fig.~\ref{fig:challenge_and_solution}(b), middle and bottom).

\subsection{Spatial and angular optimization for optical components}

During the pick-and-place of optical components, both systematic offsets and stochastic uncertainties affect the alignment of an optical setup. The latter arise from accumulated errors from the computer vision system, mechanical tolerances and uncertainties in the robotic arm motion that propagate across the workflow. The former result from human error while manually assembling the component inside the housing (a one-time task), as misalignment between the center of the housing and the center of the optical component can arise during assembly. To achieve high-precision positioning of the components comprising an optical setup, our platform employs spatial and angular optimization schemes (Fig.~\ref{fig:coarse_fine_adjustment}) designed to mitigate both systematic offsets and stochastic uncertainties. We describe the procedures below with representative experimental demonstrations.


The spatial optimization scheme operates through closed-loop feedback between the robotic arm and the beam detection camera (Fig.~\ref{fig:coarse_fine_adjustment}(a)). Aided by the other two cameras, the robotic arm places the component (e.g., a lens) at an initial position along the beam path. However, unavoidable errors cause the center of the component to deviate from the intended beam path  (by an amount $x$). This causes the beam to deviate from the target center (by an amount $y$) on the detection camera. To calibrate the system's response, the robot performs an initial perturbation by displacing the component (by $\Delta x$), which is selected to be larger than the precision of the robotic arm yet small enough to ensure the beam remains within the beam detection camera's field of view. This results in a corresponding shift in the beam position (by $\Delta y$). From the system's response to this perturbation, the system calculates the corrective movement required for the next iteration based on the first-order Newton's-method-based approach: $-\Delta x (y+\Delta y)/\Delta y$. 

The bottom two panels in Fig.~\ref{fig:coarse_fine_adjustment}(a) illustrate the spatial optimization process for two components across various initial positions. The protocol runs until the distance between the current beam position and the target position is smaller than the beam waist. Table~\ref{table:spatial_optimization} shows the results of the spatial optimization method. Both components are able to be positioned with sub-millimeter precision after the optimization.

\captionsetup[table]{labelsep=colon, skip=5pt} 

\begin{table}[t]
\centering
\caption{Results of spatial optimization}
\label{table:spatial_optimization}
\begin{tabular}{|c|c|c|c|c|}
\hline
Type & \makecell{Initial position\\ range (\unit{\mm})} & \makecell{Component placement\\ precision (\unit{\mm})} & \makecell{Beam center \\ precision (\unit{\mm})}\\ \hline \hline

\makecell{OC} & -5 to 5 & $\pm0.28$ & $\pm0.14$ \\ \hline

\makecell{Lens} & -1.25 to 0.75 & $\pm0.074$ & $\pm0.31$ \\ \hline

\end{tabular}
\end{table}

\begin{figure*}[htbp]
\centering
\includegraphics[width=0.925\textwidth]{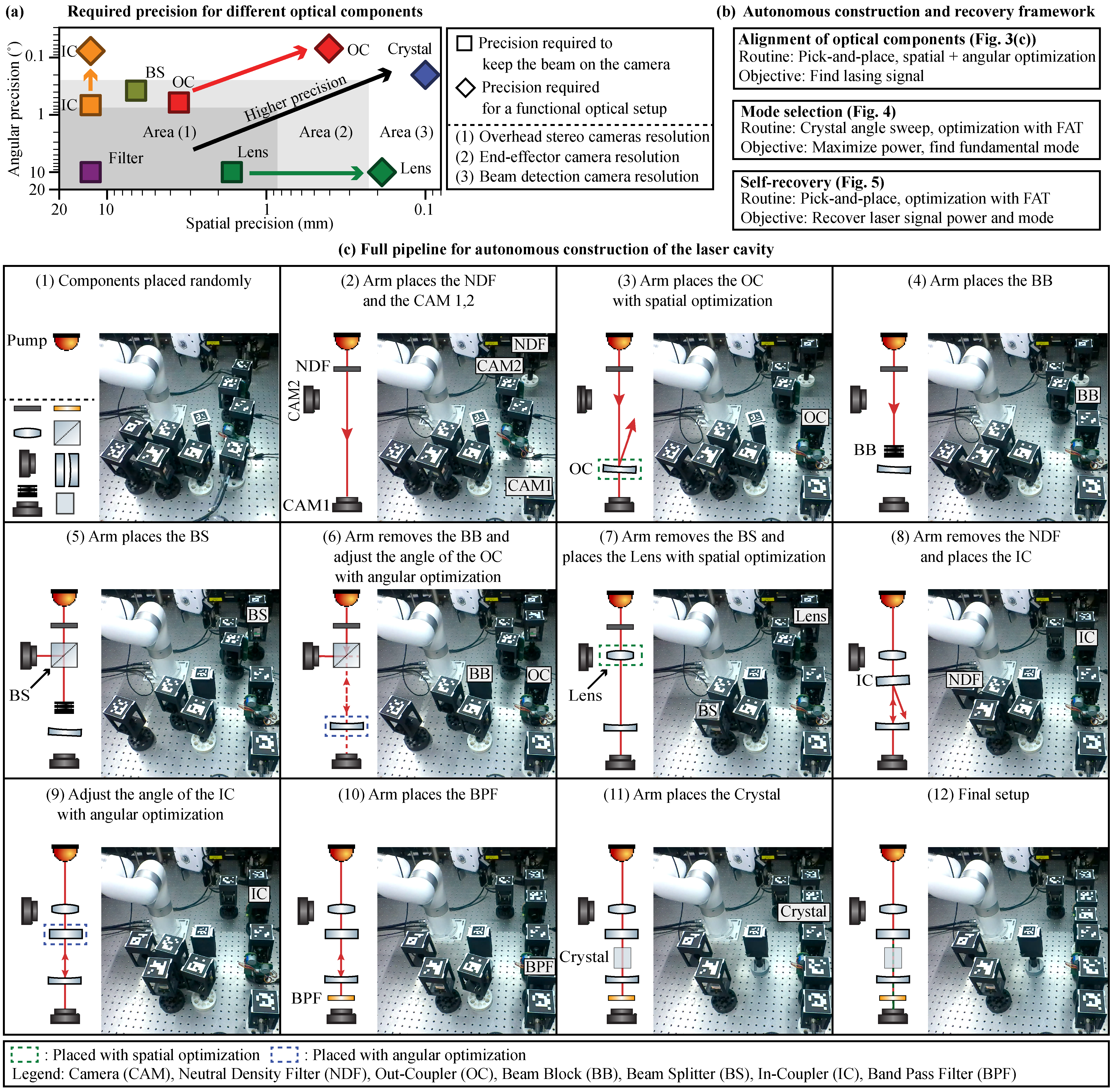}
\caption{\textbf{(a)} The spatial and angular precision necessary for the placement of various optical components for this setup. The precision for each component is calculated using the actual components used during the experiment. \textbf{(b)} Autonomous construction and recovery framework for the laser cavity. \textbf{(c)} Pipeline for autonomous construction of the setup. Videos of the steps in two distinct successful trials can be found in the project link.}
\label{fig:pipeline}
\end{figure*}

The angular optimization process is described in Fig.~\ref{fig:coarse_fine_adjustment}(b), using an example task of aligning an optical resonator. An optical resonator consists of two mirrors and its alignment is achieved by spatially overlapping two different beams: the primary beam, which is partially transmitted through both the in-coupler (IC) and the out-coupler (OC) mirrors (larger beam in Fig.~\ref{fig:coarse_fine_adjustment}(b)) and the secondary beam generated by the internal reflections between two mirrors (the smaller beam in Fig.~\ref{fig:coarse_fine_adjustment}(b)). To isolate the secondary beam, our platform stores a reference image containing only the primary beam, captured before the placement of the IC. This reference image is subtracted from the live camera feed, enabling the system to locate the secondary beam during the optimization. A Bayesian optimization algorithm is employed to minimize the distance between the centroids of the two beams. During each iteration, the FAT rotates the knobs controlling the mirror's angular position, and the distance between the two beams is updated to determine the rotation angle for each motor in the next iteration. 

As shown in Fig.~\ref{fig:coarse_fine_adjustment}(b), the beams are perfectly aligned after optimization, exhibiting an interference pattern due to the high spatial overlap between the two beams. Starting with ten different initial positions of the secondary beam, the angular optimization process succeeded in overlapping the two beams with a 100\% success rate within $5.5\pm0.5$ iterations. A similar optimization process applies when adjusting the orientation of the OC, which uses a beam splitter (BS) to collect both the reference image and the subsequent image with both beams. Importantly, our method is found to work exceedingly well when the beams are far from perfectly Gaussian, as is often the case in practical optics tasks.

\begin{figure}[htbp]
\centering
\includegraphics[width=0.475\textwidth]{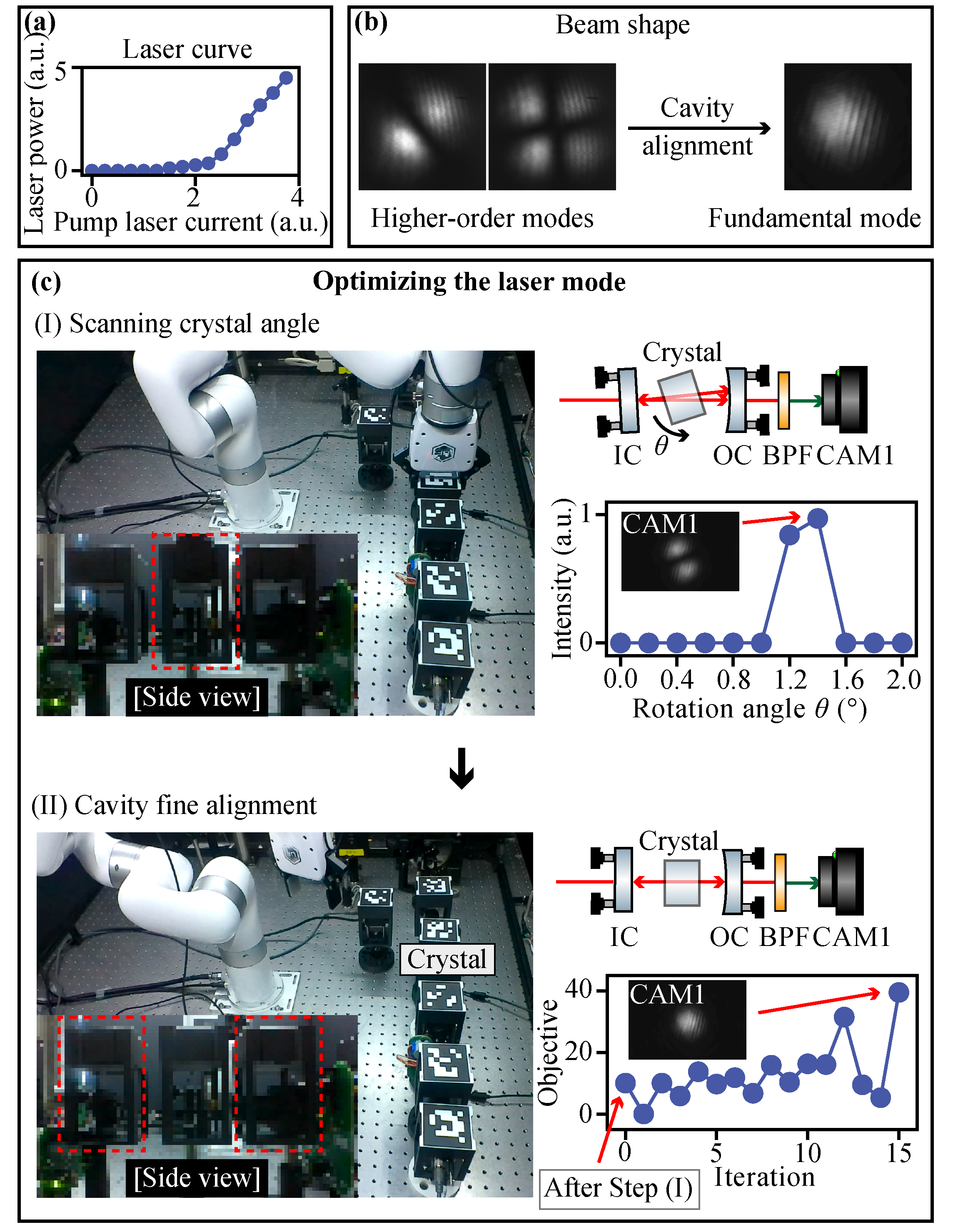}
\caption{\textbf{(a)} Observed laser power curve showing a clear threshold. \textbf{(b)} Lasing in higher-order cavity modes due to small misalignments. \textbf{(c)} Optimizing the laser mode by utilizing the robotic arm in conjunction with the FAT.}
\label{fig:crystal_rotation}
\end{figure}

\begin{figure*}[htbp]
\centering
\includegraphics[width=0.95\textwidth]{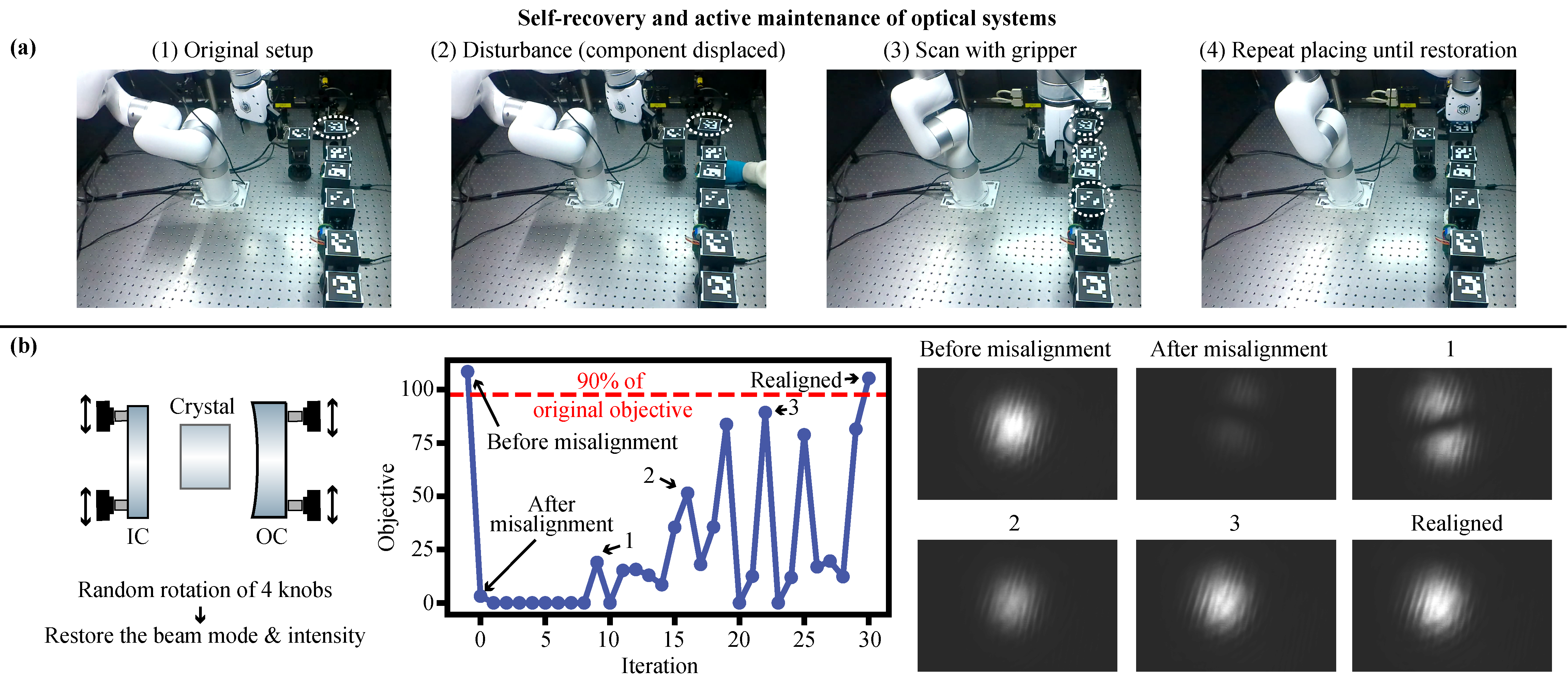}
\caption{Self-recovery of the experimental setup in response to $\textbf{(a)}$ component displacement and $\textbf{(b)}$ simulated drift of the mirror adjustment knobs.}
\label{fig:surveillance}
\end{figure*}

\section{Results}
Leveraging the full capabilities of our platform, we now demonstrate a comprehensive pipeline for the fully autonomous construction of a technically demanding optical setup: an optically pumped tabletop laser cavity (Fig.~\ref{fig:coarse_fine_adjustment}(c)). This setup embodies all challenges faced in free-space optics systems and therefore serves as a representative benchmark for robotic automation in this domain. Fig.~\ref{fig:coarse_fine_adjustment}(c) also shows the diversity of optimization objectives utilized during assembly and alignment (cross reference numbers with Fig.~\ref{fig:challenge_and_solution}(a)). The laser cavity assembly is initiated by centering the beam detection camera approximately along the intended beam path (objective 1). Lasing occurs when the pump beam (red) and the laser beam (green) generated inside the resonator are aligned (objective 2). To achieve this, the optical axis of the lens that focuses the pump beam inside the crystal must be aligned with the pump beam (objective 3), while the resonator must be precisely tuned to allow multiple round-trip interactions between the laser beam and the crystal (objective 4). Once the laser signal is observed by the camera, component parameters must be adjusted to increase the signal intensity and select a specific mode (objectives 5 and 6).

Fig.~\ref{fig:pipeline}(a) quantifies the varied precision requirements across optical components for this task. For auxiliary components (e.g., filter or BS), their functionality is maintained as long as the beam incident upon them remains within the beam detection camera's field of view (criterion 1, square markers). These tolerances fall within the resolution of the stereo cameras (dark gray area) and end-effector camera (light gray area), allowing standard vision-guided pick-and-place for their alignment. In contrast, laser cavity components are highly sensitive to small beam displacements during alignment (criterion 2, diamond markers). Applying this stricter precision criterion requires optimization based on feedback from the beam detection cameras (white area).

\subsection{Comprehensive pipeline for building laser cavity}
Prior to the laser cavity assembly, the beam path of the pump laser within the laboratory coordinate system must be measured to serve as a reference. To do this, the robotic arm grips the beam detection camera and places it at multiple in-plane coordinates (i.e., $(x,y)$ coordinates). At each coordinate, the position of the robotic arm is adjusted to center the beam on the beam detection camera. Based on these measurements, a linear fit is performed to obtain the laser beam path, which is used to determine the initial positions of each component of the optical setup defined within the laboratory coordinate system.

Once the user provides the experimental layout, the platform constructs the laser cavity without any human intervention, performing all required feedback and optimization steps (Fig.~\ref{fig:pipeline}(b)). The construction process begins with ten optical components randomly placed on the optical table by the user (Fig.~\ref{fig:pipeline}(c)(1)). The system converts the experimental layout provided by the user (e.g., distance among optical components) to the laboratory coordinates. Then two feedback detection cameras (CAM1, CAM2) and a neutral density filter (NDF) are positioned by the pick-and-place operation (Fig.~\ref{fig:pipeline}(c)(2)). The NDF is required to avoid camera saturation during the alignment. The system receives the image from the CAM1 to determine the reference coordinates of the pump beam. Then the robotic arm picks and places the OC and executes the spatial optimization routine to ensure the distance between the optimized beam and the reference beam remains smaller than the pump beam waist (Fig.~\ref{fig:pipeline}(c)(3)). Before the angular alignment of the OC, a beam block (BB) is introduced between the NDF and the OC (Fig.~\ref{fig:pipeline}(c)(4)), such that the CAM2 only detects the primary beam originating directly from the pump after placing the BS (Fig.~\ref{fig:pipeline}(c)(5)). To support an efficient angular optimization of the OC at the later stage, the position of the BS is adjusted so that the primary beam can hit near the center of the CAM2. After the BB is removed, the CAM2 detects both the primary beam and the secondary beam reflected from the OC (Fig.~\ref{fig:pipeline}(c)(6)); the system then utilizes the reference beam measured in Step 5 to run the angular optimization algorithm. Following this alignment, the robotic arm removes the BS and places the Lens (Fig.~\ref{fig:pipeline}(c)(7)) to focus the pump beam at the point where the crystal will be placed in a subsequent stage. To ensure maximum spatial overlap with the laser mode, the position of the Lens is adjusted via the spatial optimization. The IC is subsequently placed (Fig.~\ref{fig:pipeline}(c)(8)) and its angular orientation is optimized to form an optical resonator (Fig.~\ref{fig:pipeline}(c)(9)). A band pass filter (BPF) is placed to isolate the laser signal from the pump and only observe the generated laser signal with the CAM1 (Fig.~\ref{fig:pipeline}(c)(10)), followed by the placement of the Nd:YAG crystal (Fig.~\ref{fig:pipeline}(c)(11)). The position of the crystal and the cavity parameters (i.e., the orientation of the IC and the OC controlled by the FAT) can be tuned to maximize the laser signal and optimize the laser mode, as described in the following section. The power curve showing the expected threshold behavior is given in Fig.~\ref{fig:crystal_rotation}(a), indicating that successful lasing is observed in the final setup (Fig.~\ref{fig:pipeline}(c)(12)).

\subsection{Mode selection and maximizing output power}

Besides the optical resonator geometry, the angular orientation of the crystal acts as yet another tuning parameter that determines the power and shape of the lasing mode. When the resonator and the crystal are aligned properly, the system favors the fundamental mode, which exhibits a Gaussian intensity profile, as demonstrated in the right panel of Fig.~\ref{fig:crystal_rotation}(b). Conversely, misalignment of the crystal and/or resonator promotes the amplification of higher-order modes, which are visualized in the left panel of Fig.~\ref{fig:crystal_rotation}(b). The proximity of the laser mode to the fundamental mode is quantified by the beam quality factor $M^2$, where $M^2=1$ denotes a diffraction-limited Gaussian beam and higher values indicate the existence of higher-order modes in the beam profile. Therefore, the primary objective here is to achieve the fundamental mode with maximum signal intensity. 

Fig.~\ref{fig:crystal_rotation}(c) demonstrates the protocol for finely adjusting the laser cavity to maximize output power in the fundamental mode. This is done using the combined action of the robotic arm and the FAT. Initially, one of the knobs for the IC was rotated to introduce misalignment in the resonator. As the first step for optimization, the robotic arm performs a coarse angular sweep of the crystal in increments of 0.2\textdegree, with the laser beam image recorded via the CAM1 at each step (Fig.~\ref{fig:crystal_rotation}(c)). Due to an initial misalignment, the laser signal is observed only within a narrow angular range ($\theta = 1.2 ^\circ \sim 1.4^\circ$) and in a higher-order mode. Once the robotic arm identifies the optimal orientation of the crystal, the end-effector is released and Bayesian optimization is executed to adjust the cavity mirrors via the FAT. The objective function is defined as the ratio of the square root of the beam intensity $I$ to the beam quality factor $M^2$ : $\sqrt{I}/M^2$. Following this optimization, the laser cavity supports the fundamental mode.

\subsection{Self-recovery under external perturbations}
Autonomous operation in optical laboratories requires not only initial assembly and alignment, but also the ability to maintain functionality under external perturbations. In contrast to the manual and time-intensive recovery procedures in conventional optics experiments, we implement a self-recovery procedure based on continuously logged component configurations and system responses which aims to provide long-term stability and reproducibility of optics experiments.

In this section, we describe how our platform facilitates the self-recovery of perturbed optical setups through continuous monitoring. Fig.~\ref{fig:surveillance} provides proof-of-concept experiments for setup reconstruction under two catastrophic scenarios. In Fig.~\ref{fig:surveillance}(a), we simulate an accidental, human-induced displacement of the lens ($\sim$ \unit{\cm} scale). To initiate reconstruction, the end-effector camera measures the current positions of the critical optical elements that affect the performance of the laser cavity (lens, IC, and OC). By comparing the current positions with the original positions stored before the misalignment, the system can identify the misplaced component and execute corrective action.

Because the finite precision of the robotic arm movement leads to imperfect recovery with a one-time pick-and-place attempt, the system employs a realignment routine after moving the component from the displaced position to the original position. The realignment routine is activated if the laser signal remains absent after the initial pick-and-place action and continues until the laser signal is restored. During this routine, the robotic arm grips the component, translates it, and repositions it at the same in-plane coordinates. The complete reconstruction pipeline (step (1) to (4) in Fig.~\ref{fig:surveillance}(a)) is validated across 10 independent trials, with its result shown in Table~\ref{table:displacement_recovery}. The system restored the laser signal in all trials (100\% success); 40\% of trials achieved restoration with a single pick-and-place (i.e., without the realignment routine). The remaining cases could also recover the signal with $2.7\pm1.4$ pick-and-place attempts during the realignment routine. The average recovery time is similar to the time required for the spatial optimization of the lens in Fig.~\ref{fig:pipeline}(c)(7), showing that the recovery can be accomplished within a similar timescale to a single step during a full-laser construction.

\captionsetup[table]{labelsep=colon, skip=5pt} 

\begin{table}[t]
\centering
\caption{Results of displacement recovery}
\label{table:displacement_recovery}
\begin{tabular}{|c|c|c|c|c|}
\hline
\makecell{Success \\ rate} &  \makecell{Attempts during \\ realignment routine} & \makecell{Average\\ recovery time} & \makecell{Lens optimization \\ time in Fig.~\ref{fig:pipeline}(c)(7)}\\ \hline \hline

10/10 & $2.7\pm1.4$  & \SI{2.83}{min} & \SI{2.6}{min}  \\ \hline
\end{tabular}
\end{table}

\captionsetup[table]{labelsep=colon, skip=5pt} 

\begin{table}[t]
\centering
\caption{Results of environmental drift recovery}
\label{table:environmental drift recovery}
\begin{tabular}{|c|c|c|c|c|}
\hline
\makecell{Success \\ rate} &  \makecell{\# of Iterations \\during recovery} & \makecell{Average\\ recovery time}& \makecell{Resonator alignment\\time in Fig.~\ref{fig:pipeline}(c)(9)}\\ \hline \hline

9/10 & $12\pm9$  & \SI{3.05}{min} & \SI{2.83}{min} \\ \hline

\end{tabular}
\end{table}

Fig.~\ref{fig:surveillance}(b) demonstrates the reconstruction process following a large misalignment of the optical resonator, e.g., due to a stochastic perturbation from the environment. This is induced by rotating all four mirror knobs randomly between 30\textdegree~and 60\textdegree~in arbitrary directions. This perturbation causes a loss of lasing, resulting in no detected signal. To restore the cavity alignment, we run a Bayesian optimization routine similar to the one demonstrated in Fig.~\ref{fig:crystal_rotation}. The objective function is set to $I/M^2$, and the convergence threshold is defined as 90\% of the initial objective function to facilitate convergence, accounting for the possibility that the system initially started from the global optimum. Table~\ref{table:environmental drift recovery} demonstrates the experimental results of testing the realignment of the laser cavity starting from 10 different random knob positions. In most cases, the system could restore the laser signal on a timescale similar to that required for the cavity alignment during the initial construction (Fig.~\ref{fig:pipeline}(c)(9)). The middle and right panels in Fig.~\ref{fig:surveillance}(b) demonstrate one of the attempts. The laser beam recovers its Gaussian shape while the beam intensity approaches that of the original beam.

Together, these results demonstrate that our platform is capable of persistent maintenance of complex optical experiments, enabling robust long-term autonomous operation.

\section{CONCLUSIONS}
We have developed a robotics framework capable of autonomously constructing, aligning, and maintaining precision free-space optical setups, and demonstrated it to realize a tabletop laser cavity from scratch. Starting from randomly distributed components, the system performs assembly, optical alignment, mode selection, and self-recovery from disturbances without human intervention. This result establishes that highly complex optical experiments, traditionally dependent on manual expertise, can be reformulated as feedback-driven autonomous workflows. In doing so, this framework removes the last remaining hurdle for scalable laboratory automation in optics. More broadly, our work points towards laboratories in which experimental configurations are no longer static and manual, but are dynamically controlled systems capable of sustained operation and self-correction.

Looking forward, several directions could further advance our platform. Incorporating predictive optical simulations, agentic AI or learning-based strategies into the workflow could further simplify the design and execution of optical experiments and improve performance. In parallel, integration with remote interfaces to create ``cloud-laboratories" would enable democratized access to complex optical instrumentation. Finally,  continued hardware improvements, such as wireless charging for electronic devices (e.g., FAT, beam detection cameras) on the optical table and a reduced form factor for the 3D-printed housings, would expand the scale of optical experiments that our platform can automate.

\addtolength{\textheight}{-12cm}   





\section*{ACKNOWLEDGMENT}

The authors thank Ryan Lopez, Sahil Pontula, Serena Landers, David Dai, and Hae Won Lee for stimulating discussions through this project. S.C. acknowledges support from the Korea Foundation for Advanced Studies Overseas PhD Scholarship. S.V. and M.S. acknowledge support from the National Science Foundation under Cooperative Agreement PHY-2019786 (The NSF AI Institute for Artificial Intelligence and Fundamental Interactions, \href{http://iaifi.org/}{http://iaifi.org/}). This material is based upon work supported in part by the U.S. Army DEVCOM ARL Army Research Office through the MIT Institute for Soldier Nanotechnologies under Cooperative Agreement number W911NF-23-2-0121. We also acknowledge the support of Parviz Tayebati, the MIT Undergraduate Research Opportunities Program (UROP), and the MIT Generative AI Impact Consortium (MGAIC).


\end{document}